\documentclass[letterpaper, 10 pt, conference]{ieeeconf}  

\IEEEoverridecommandlockouts                              

\usepackage{graphics} 
\usepackage{epsfig} 
\usepackage{times} 
\usepackage{amsmath} 
\usepackage{amssymb}  
\usepackage{bbm}

\usepackage{multicol}
\usepackage{graphicx}
\usepackage{bm,amsfonts}
\usepackage{subfigure}
\usepackage[dvipsnames]{xcolor}

\makeatletter
\let\NAT@parse\undefined
\makeatother
\usepackage{hyperref}
\hypersetup{
    colorlinks=true,
    linkcolor=orange,
    filecolor=magenta,      
    urlcolor=orange,
    citecolor=orange,
}

\title{\LARGE \bf Leveraging Smooth Attention Prior\\for Multi-Agent Trajectory Prediction}

\author{Zhangjie Cao$^1$, Erdem B\i y\i k$^2$, Guy Rosman$^3$, Dorsa Sadigh$^1$%
    \thanks{Emails: {\tt\footnotesize caozj@cs.stanford.edu, ebiyik@stanford.edu, guy.rosman@tri.global, dorsa@cs.stanford.edu}}%
    \thanks{$^1$Computer Science, Stanford University, CA, USA}%
    \thanks{$^2$Electrical Engineering, Stanford University, CA, USA}%
    \thanks{$^3$Toyota Research Institute, MA, USA}%
    }

\begin{document}

\maketitle
\thispagestyle{empty}
\pagestyle{empty}

\begin{abstract}
Multi-agent interactions are important to model for forecasting other agents' behaviors and trajectories. At a certain time, to forecast a reasonable future trajectory, each agent needs to pay attention to the interactions with only a small group of most relevant agents instead of unnecessarily paying attention to all the other agents. However, existing attention modeling works ignore that human attention in driving does not change rapidly, and may introduce fluctuating attention across time steps. In this paper, we formulate an attention model for multi-agent interactions based on a total variation temporal smoothness prior and propose a trajectory prediction architecture that leverages the knowledge of these attended interactions. We demonstrate how the total variation attention prior along with the new sequence prediction loss terms leads to smoother attention and more sample-efficient learning of multi-agent trajectory prediction, and show its advantages in terms of prediction accuracy by comparing it with the state-of-the-art approaches on both synthetic and naturalistic driving data. We demonstrate the performance of our algorithm for trajectory prediction on the INTERACTION dataset on our \href{https://sites.google.com/view/smoothness-attention}{\color{orange}{website}}\footnote{https://sites.google.com/view/smoothness-attention}.
\end{abstract}

\section{Introduction}\label{sec:intro}
To navigate safely and efficiently in dense and complex traffic scenarios crowded by vehicles and pedestrians, it is crucial for autonomous vehicles or mobile robots to be socially compliant. This requires interacting with other agents and making decisions based on not only the observation of the environment, but also the behaviors of other agents. For example, you might decide to slow down if you see a vehicle aggressively taking over another vehicle in your rear view mirror. Hence, socially compliant navigation requires accurately forecasting future trajectories of the surrounding agents~\cite{lefevre2014survey,schwarting2019social}. However, this trajectory prediction problem can be challenging in complex traffic scenarios, as many agents might be interacting with each other. 

Taking into account the interaction with all other agents for the goal of predicting an agent's trajectory can be computationally intractable
\textemdash such an approach may require too many training trajectories, far beyond the size of existing datasets~\cite{chang2019argoverse,houston2020one}. While part of this problem is remedied by parameter sharing between agents~\cite{bhattacharyya2018multi}, an important observation is that humans can effectively coordinate with each other in such complex driving scenarios without reacting to \emph{all other agents} at the same time \cite{scalf2013competition}. 
People attend only to a limited set of factors at every given moment~\cite{treisman1969modelsofattention}. In the driving context, the factors are mostly confined to a subset of agents and objects that immediately affect or are affected by them. For example, when driving in a lane, the driver often pays attention only to the vehicles in neighboring lanes with potential to join the lane. Such limited attention largely reduces the complexity of the model and the amount of training data needed~\cite{vemula2018social}. 

Many existing works learn attention models that are purely based on the goal of trajectory prediction~\cite{vemula2018social,gupta2018social}. However, prior knowledge can substantially improve the multi-agent trajectory prediction~\cite{casas2020importance}. Attention priors such as the neighboring prior that agents only pay attention to nearby agents~\cite{alahi2016social}, and the sparsity prior that the attention should only be paid to a limited number of agents~\cite{zhang2018attention} are helpful in trajectory prediction in complex traffic scenes. However, the neighboring prior is often easily violated since there are situations where not just nearby agents affect the ego agent's behavior, \emph{e.g.,} vehicles need to pay attention to an ambulance and make way for it even when it is far away. The sparsity prior can also make errors when the agents attend to limited but the incorrect agents. In this paper, we provide a new attention prior that is motivated by cognitive science and is widely applicable to different driving scenarios.

We notice that changes in a traffic scenario are usually not fast enough to require rapid changes in the attention, and cognitive science literature suggests humans' social attention does not change rapidly~\cite{swettenham1998frequency}. Thus, we propose a temporal smoothness prior on the attention model. Specifically, we train an attention model with a constraint that the attention distribution over consecutive time steps does not change rapidly. 
Using this attention regularization for smoothness, we design a network architecture to simultaneously model interactions among agents and learn the attention of each agent to all other agents, where we impose the smooth attention prior as a total variation loss on the attention. We then perform trajectory prediction for each agent by a recurrent neural network with the history of states and the predicted attention as input.

Our contributions in this paper are three-fold:
\begin{itemize}
    \item We propose a smoothness prior on attention modeling by constraining its rate of change. This prior is based on evidence for smooth attention from cognitive science and the commonly smooth nature of traffic scenarios.
    \item We develop an architecture that simultaneously learns an attention model for interactions and a sequential prediction model for predicting future states. The architecture is optimized by a sequence likelihood loss to minimize prediction error and a total variation loss to impose the smoothness attention prior.
    \item We conduct experiments on synthetic and real multi-agent trajectory prediction datasets. Our results show the proposed approach consistently achieves the lowest prediction errors and increases sample-efficiency compared to the state-of-the-art trajectory prediction methods.
\end{itemize}

\section{Related Work}
\noindent\textbf{Trajectory Prediction.}
Multi-agent trajectory prediction aims to predict the future trajectories of a group of interactive agents given their observation history. Helbing et al. introduced \emph{social force} that includes attractive and repulsive forces between agents to model the behavior of pedestrians~\cite{helbing1995social}.
More recently, different sequential prediction models have been proposed to enable trajectory prediction including Gaussian processes~\cite{kim2011gaussian}, hidden Markov models~\cite{li2019generic}, dynamic Bayesian networks~\cite{kasper2012object}, inverse reinforcement learning~\cite{sun2018probabilistic}, and recurrent neural networks (RNN)~\cite{vemula2018social,alahi2016social,gupta2018social,chandra2020forecasting,huang2020diversitygan,pokle2019deep}. However, these methods do not employ a separate module to explicitly reason about interactions between the agents, but implicitly model the interactions as part of the trajectory prediction problem.

\noindent\textbf{Multi-Agent Interactions.}
Modeling interactions in multi-agent systems is important not only for autonomous driving, but more generally for robotics. Many works handle interactions by learning and modeling the other agents in the environment~\cite{basu2019active,kwon2020when,xie2020learning,zhu2020multi,sadigh2016information,sadigh2016planning,sadigh2018planning}, and Communication protocols and conventions are developed to enable effective interactions between agents~\cite{sukhbaatar2016learning,shih2021critical}. Furthermore, 
graph structures are adopted to model interactions~\cite{kipf2018neural,bohmer2020deep}. In driving, interactions between agents usually have common structures, \emph{e.g.,} agents in the neighborhood are more likely to interact with each other. Therefore, driving prediction techniques have an opportunity to leverage these structures for better performance.

\noindent\textbf{Trajectory Prediction via Interaction Modeling.} Modeling the interactions between agents is important for multi-agent trajectory prediction since the trajectory of each agent depends on not only their internal states but also their interactions with other agents. 
Social pooling \cite{alahi2016social,gupta2018social} and graph neural networks~\cite{casas2020implicit,salzmann2020trajectron++} are adopted to simultaneously model interaction and conduct trajectory prediction. However, not all interactions between agents are necessary for trajectory prediction, and hence attention models are employed to only focus on \emph{important} agents. \emph{Social attention}~\cite{vemula2018social} designs a attention module on a graph structure to simultaneously predict attention and model interactions. EvolveGraph~\cite{li2020evolvegraph} adaptively evolves the latent interaction graphs to enable dynamic relational reasoning. Multi-head attention social pooling models the high-order interactions between vehicles~\cite{messaoud2020attention}.

While these works attempt to focus the attention on the correct agents in various ways, they lack a natural model of human attention, and do not explicitly use the properties of driving. We hypothesize that additional priors can enable more effective trajectory prediction in complex multi-agent scenarios. We aim to fill this research gap by incorporating a smooth attention prior in trajectory prediction.

\noindent\textbf{Attention Regularization.} 
Attention mechanisms aim to re-distribute the focus of the network on inputs such as agents~\cite{vemula2018social} or image parts~\cite{bahdanau2015neural}. Early works on attention focus on the attention structure and implicitly learn the attention with the task loss~\cite{bahdanau2015neural,luong2015effective}. However, implicit attention learning may result in low-quality or even meaningless attention such as too dense values~\cite{luong2015effective}. Instead, regularization techniques such as sparsity encourage the network to focus on a small number of inputs~\cite{martins2020sparse}. 
Consistency regularizer enables coherent attention at different levels of the network~\cite{zhou2019discriminative}. The sparse and structured neural attention utilizes the structural properties of the input~\cite{niculae2017regularized}. Diversity regularization maximizes attention difference between different heads of neural network~\cite{li-etal-2018-multi-head}. However, most of these attention regularizers are designed for general attention modeling in various applications, but not specially for the multi-agent trajectory prediction in complex traffic scenarios, where accurately modeling humans' attention becomes crucial. In this work, backed by evidence from cognitive science, we use the total variation attention regularizer for modeling realistic driving behavior.

\section{Method}
We are interested in the multi-agent trajectory prediction problem in complex driving scenes.
More formally, observing the states of all agents from time step $1$ to $T_\text{obs}$, we aim to predict their future states from time step $T_\text{obs}+1$ to $T_{\text{pred}}$. 

As discussed in Section~\ref{sec:intro}, trajectory prediction for each agent in the scene requires modeling the interactions between agents, which we achieve by modeling the attention of each agent to the most relevant agents. To this end, we build our architecture based on the RNN mixture model of~\cite{vemula2018social}. 
The main difference is that we propose a new smoothness attention prior on the attention model, which is implemented as a total variation prior on learning the attention.

\begin{figure}
    \centering
    \includegraphics[width=\columnwidth]{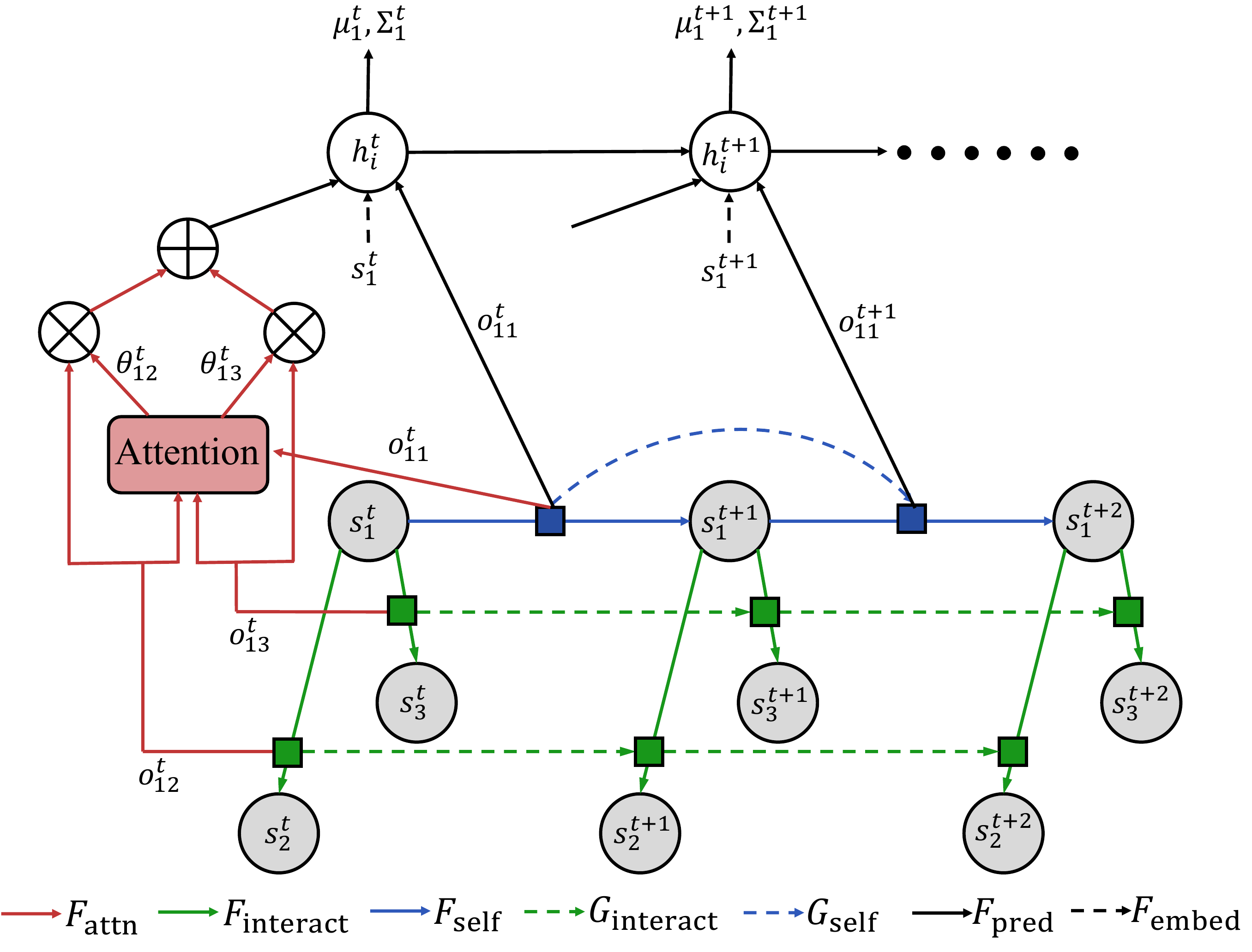}
    \vspace{-20pt}
    \caption{The architecture of the proposed method. We show the prediction process of a node (node 1) in a scene with $3$ agents at time step $t$ and $t+1$. The solid green edges indicate the interaction between different agents with the states of the connected nodes as the input. The dashed green arrows show the RNN for the solid green edges. The solid blue edges indicate the interaction between consecutive steps with the states of consecutive steps as the input. The dashed blue arrows show the RNN for the solid blue edges. The solid red arrows indicate the attention module. The solid black arrows show the input and output of the RNN to predict the location of node 1.}
    \label{fig:arch}
    \vspace{-15pt}
\end{figure}

\smallskip
\noindent \textbf{Architecture.}\label{sec:arch}
Fig.~\ref{fig:arch} shows our network architecture at time step $t$ and $t+1$ for an agent. To predict future trajectories of each agent, we incorporate information about the past and the interactions between the agent and other agents present in the scene. 
We model the interactions between agents with a directed complete graph with self-loops, where the nodes represent the agents, the edges between different agents model the interactions between agents, and self-loops model the interactions between consecutive time steps of each agent. For the edges between different agents, we adopt a neural network $F_\text{interact}$ with the states of the edge's tail and head nodes as the input, which embeds the interaction between these two nodes. Specifically, given the state $s^t_i$ of agent $i$ at time $t$, we embed the interaction between each pair of agents $(i,j)$ as 
\begin{equation}
\begin{small}
    x^t_{ij} = F_\text{interact}(\text{concat}(s^t_i, s^t_j)).
    \end{small}
\end{equation}
where `concat' here mean the vector concatenation of two states.
The self-loop models the interaction of states between consecutive time steps. We adopt a neural network $F_\text{self}$ that embeds the temporal state change of each agent. Specifically, at each time step $t$, given the current state $s^t_{i}$ and the previous state $s^{t-1}_i$ of agent $i$, we embed the temporal change as 
\begin{equation}
\begin{small}
    x^t_{ii} = F_\text{self}(\text{concat}(s^{t-1}_i, s^t_i)).
    \end{small}
\end{equation}
Modeling the interaction for each time step independently may lose the information of the temporal evolution of interactions. Thus, we adopt two RNNs $G_\text{interact}$ and $G_\text{self}$ to model the temporal evolution of edges and self-loops in the graph, where the computation at time step $t$ can be represented as
\begin{equation}
\begin{small}
\begin{aligned}
    &o^t_{ij},h^{t}_{ij} = G_\text{interact}(h^{t-1}_{ij}, x^t_{ij})\\
    &o^t_{ii},h^{t}_{ii} = G_\text{self}(h^{t-1}_{ii}, x^t_{ii}).\\
\end{aligned}
\end{small}
\end{equation}
Here, $o$ indicates the output of the RNN, and $h$ indicates the hidden states in RNN, which are the internal states of RNN to transit information from a step to another. $o^t_{ij}$ embeds the interaction information between each pair of agents and $o^t_{ii}$ embeds the temporal shift information of each agent.

As we discussed in Section~\ref{sec:intro}, each agent should focus on the interactions with the most relevant agents. For each agent $i$, we compute its attention on other agents as the similarity between $o^t_{ii}$ and the interaction features with other agents $o^t_{ij}$, where $i\neq j$:
\begin{equation}
\begin{small}
    \theta^t_{ij}= \langle F_\text{attn}(o^t_{ii}), F_\text{attn}(o^t_{ij}) \rangle,
    \end{small}
\end{equation}
Here, we use $F_\text{attn}$ to refer to a learnable attention module and use inner-product as the similarity measurement. We apply the following normalization to make the attention from one agent to all other agents a categorical distribution:
\begin{equation}\label{eqn:attention_one_step}
\begin{small}
    \theta^t_{ij} = \frac{\exp(\langle F_\text{attn}(o^t_{ii}), F_\text{attn}(o^t_{ij}) \rangle) }{\sum_{k\ne i} \exp(\langle F_\text{attn}(o^t_{ii}), F_\text{attn}(o^t_{ik}) \rangle) }\:.
    \end{small}
\end{equation}

With the interaction features and the attention, we use another RNN, $G_\text{pred}$, to predict the future trajectories of each agent. $G_\text{pred}$ uses the attended interactions between the agent and other agents, and between consecutive steps of the agent. Similar to $F_\text{interact}$ and $F_\text{self}$, we also embed the input state into a hidden space. At time step $t$, the prediction of the state of agent $i$ can be formalized as follows:
\begin{equation}
\begin{small}
    o^{t+1}_{i},h^{t+1}_i = G_\text{pred}(h^{t-1}_{i}, F_\text{embed}(s^{t-1}_i), \sum_{j\ne i}\theta^t_{ij}o^t_{ij}, o^t_{ii}),
    \end{small}
\end{equation}
where $F_\text{embed}$ is the embedding network of the agent state. We predict the current step state $\hat{s}^t_i$ by a prediction network $F_\text{pred}$ based on the output of $G_\text{pred}$, which is represented by a mean and covariance:
\begin{equation}\label{eqn:pred_one_step}
\begin{small}
    \mu^{t+1}_{i}, \Sigma^{t+1}_{i}= F_\text{pred}(o^{t+1}_i).
    \end{small}
\end{equation}
Using the above architecture, the trajectory prediction of each agent depends on the historical sequence of states and the attended interactions with other agents. Next, we introduce the loss function to train this network.

\smallskip
\noindent \textbf{Likelihood and Variance Losses.}
We define a loss function that maximizes the likelihood of the observed states:
\begin{equation}\label{eqn:one_step_likelihood}
\begin{small}
\begin{aligned}
    L_{\textrm{likelihood}}(s,\mu,\Sigma) = -\sum_{i=1}^N\sum_{t=1}^{T_{\textrm{pred}}} \log P(s_i^t\mid \mu_i^t,\Sigma_i^t).
\end{aligned}
\end{small}
\end{equation}
where $P(s_i^t \mid \mu_i^t,\Sigma_i^t)$ is the probability of $s_i^t$ under a Gaussian distribution with mean $\mu_i^t$ and covariance matrix $\Sigma_i^t$.
Only minimizing the likelihood loss can sometime introduce large variance for the predictions, and can cause unstable predictions during test time. 
We therefore place a prior on the variance via an additional loss:
\begin{equation}\label{eqn:one_step_std}
\begin{small}
    \begin{aligned}
    L_{\textrm{var}}(\Sigma) = \sum_{i=1}^N\sum_{t=1}^{T_{\textrm{pred}}}\sum_{k=1}^{\textrm{dim}(s)}\left(\exp\left(\left({\Sigma_{i}^t}\right)_{kk}^\frac{1}{2}\right)\mathbb{I}\left(\left({\Sigma_{i}^t}\right)_{kk}^\frac{1}{2}> \tau\right)\right)
\end{aligned}
\end{small}
\end{equation}
where $\mathbb{I}(\cdot)$ is the indicator function, and threshold $\tau_i$ ensures that we do not penalize small variances.

The above losses, originally used in~\cite{vemula2018social}, only penalize the one-step error and under-estimate the compounding error in long horizon. At test time, a sequence of states that come after the observed time step are predicted, where errors will accumulate over time steps and introduce a large compounding error. To address this problem, in the training phase, we simulate the prediction process of the test phase as follows:
\begin{equation}\label{eqn:pred_sequence}
\begin{small}
    \begin{aligned}
    \hat{o}^{t+1}_{i},\hat{h}^{t+1}_i &= G_\text{pred}(\hat{h}^{t-1}_{i}, F_\text{embed}(\hat{s}^{t-1}_i), \sum_{j\ne i}\hat{\theta}^t_{ij}\hat{o}^t_{ij}, \hat{o}^t_{ii})\:,\\
    \hat{\mu}^{t+1}_{i}, \hat{\Sigma}^{t+1}_i &= F_\text{pred}(\hat{o}^t_i)\:,
\end{aligned}
\end{small}
\end{equation}
To differentiate from the notation above for one-step prediction, we use the notation with $\hat{\cdot}$ to indicate that the variable is derived using the multi-step prediction process.  
$\hat{s}^{t}_i$ is a sample from a Gaussian distribution parameterized by $\hat{\mu}^{t}_i$ and $\hat{\Sigma}^{t}_i$ if $t>T_\text{obs}$ or otherwise it is the ground-truth state. We similarly apply the Gaussian likelihood loss $L_{\textrm{likelihood}}(\hat{s},\hat{\mu},\hat{\Sigma})$ and the variance loss $L_{\textrm{var}}(\hat{\Sigma})$ on this simulated test-time prediction, which
are defined on the full prediction sequence, in contrast to the one-step loss in Eqns.~\eqref{eqn:one_step_likelihood} and~\eqref{eqn:one_step_std}. With the above loss functions, we can simultaneously penalize the short-term and long-term prediction errors.

The attention module, which is used as an input to the RNN to predict $\mu$ and $\Sigma$, is implicitly learned by the above losses without any explicit constraints. However, proper regularization might be helpful in training an attention model, which is effective in reducing the sample complexity. Therefore, we further introduce a regularization on the attention to learn a more reasonable and informative attention module.

\smallskip
\noindent \textbf{Attention Modeling.}\label{sec:attention}
In this section, we impose the smooth attention prior that human attention does not change frequently over time. Our hypothesis is based on two factors: First, cognitive science literature suggests that although human attention can change rapidly when running freely without specific instructions, deliberate movement of attention is significantly slower because of an internal limit on the speed of volitional commands~\cite{wolfe2000attention}. This suggests that human social attention does not change frequently in driving, as it falls into the category of deliberate movement. Second, in driving scenarios, the most relevant vehicles to pay attention to are often the ones that can immediately affect the ego agent or the ones that are affected by the ego agent. This group of agents often do not change their behavior rapidly since the reward function typically consists of terms related to the distance to the goal and proximity to other agents, which mostly continuously change along time steps. Hence, we impose a smoothness constraint on the attention, which is defined as a vectorial total variation penalty: 
\begin{equation}\label{eqn:smooth}
\begin{small}
\begin{aligned}
    &L_{\textrm{smooth}}(\theta) = \sum_{t=2}^{T_{\textrm{pred}}}\sum_{i=1}^N \sqrt{\sum_j (\theta_{ij}^t - \theta_{ij}^{t-1})^2\:},\\
\end{aligned}
\end{small}
\end{equation}
This total variation loss \cite{rudin1992nonlinear,bresson2008fast} encourages piecewise-smooth signals with few transitions, and transitions that happen at the same point for multiple elements of $\theta_t$. We similarly define the total variation loss on $\hat{\theta}$. 

\noindent\textbf{Overall Loss.} We aggregate all the losses for the network:
\begin{equation}
\begin{small}
\begin{aligned}
    L = &L_{\textrm{likelihood}}(s,\mu,\Sigma) + L_{\textrm{likelihood}}(\hat{s},\hat{\mu},\hat{\Sigma}) + \beta_1( L_{\textrm{var}}(\Sigma) + L_{\textrm{var}}(\hat{\Sigma})) \nonumber\\
    &+ \beta_2 (L_{\textrm{smooth}}(\theta)+L_{\textrm{smooth}}(\hat{\theta}))\:,\label{eqn:loss_whole}
\end{aligned}
\end{small}    
\end{equation}
where $\beta_1$ and $\beta_2$ are trade-offs for the variance losses and the smoothness losses. $s$, $\mu$, $\Sigma$, and $\theta$ correspond to one-step predictions as shown in Eqns.~\eqref{eqn:attention_one_step} and~\eqref{eqn:pred_one_step}. $\hat{s}$, $\hat\mu$, $\hat\Sigma$, and $\hat\theta$ are sequence predictions  as shown in Eqn.~\eqref{eqn:pred_sequence}.

\section{Experiments}
We first introduce the experiment setting including datasets, and implementation details. We then show our results where we perform trajectory (state) prediction.

\subsection{Datasets}
We perform tests on two synthetic datasets: Double Merge and Halting Car, for which we implement the scenarios in the CARLO simulator~\cite{cao2020reinforcement}. The synthetic experiments aim to show the attention learned by our smoothness prior focuses on the correct agents, where the scenarios usually exists on the road and the true attention is clear. We also conduct experiments on the real naturalistic driving INTERACTION dataset~\cite{interactiondataset}.

\begin{figure}[ht]
    \centering
     \includegraphics[width=.49\textwidth]{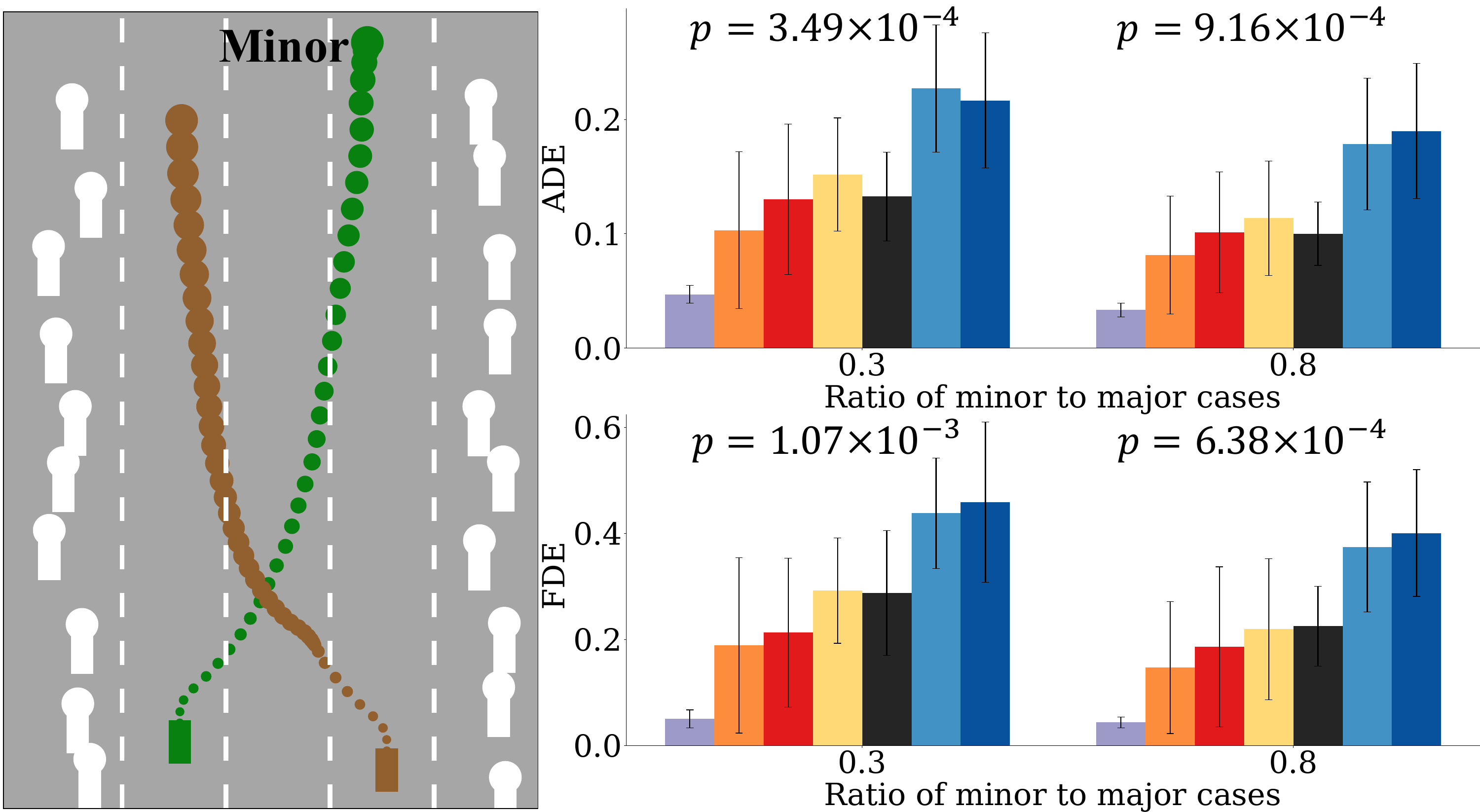}\label{fig:minor}
    \includegraphics[width=.49\textwidth]{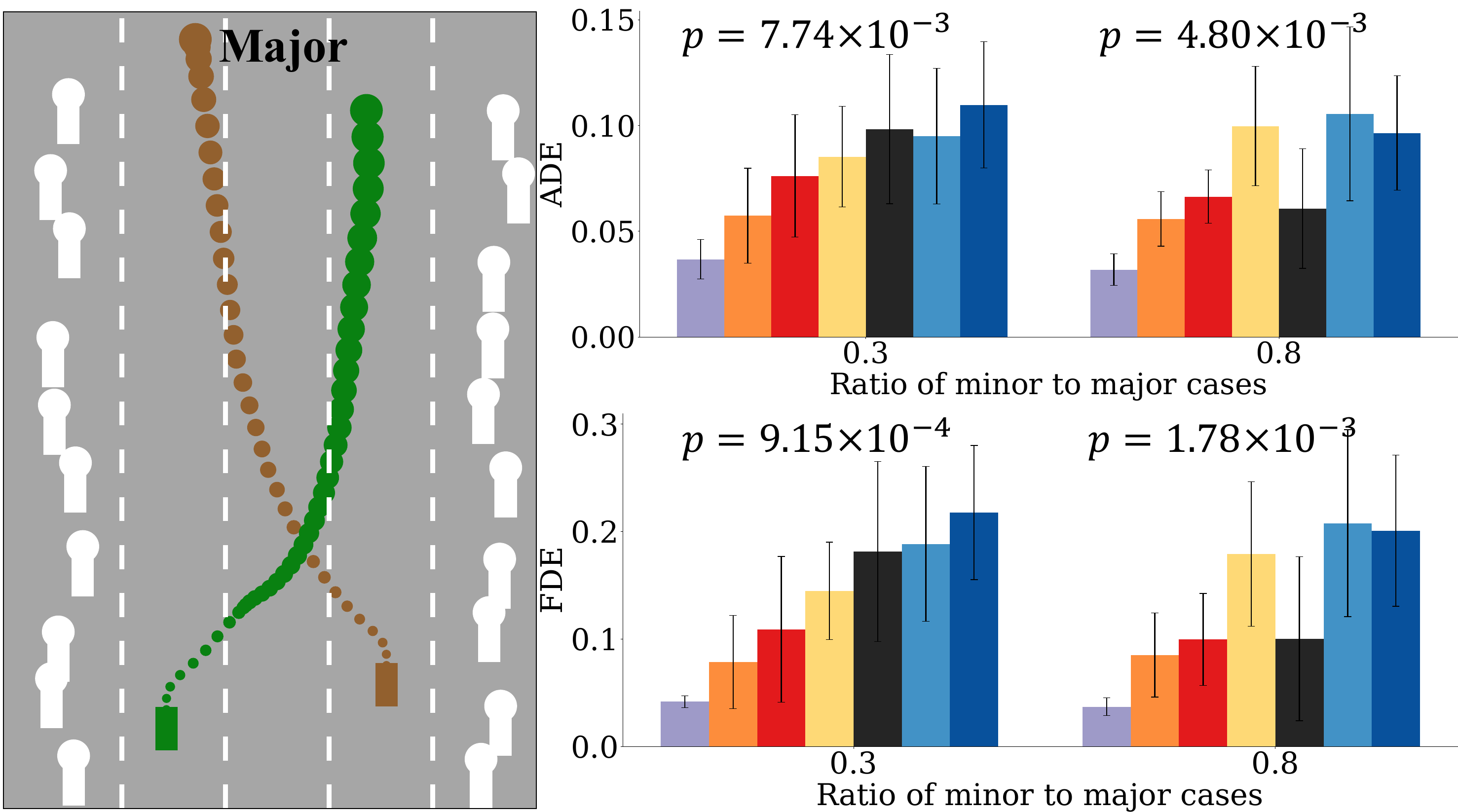}\label{fig:major}
    \includegraphics[width=\columnwidth]{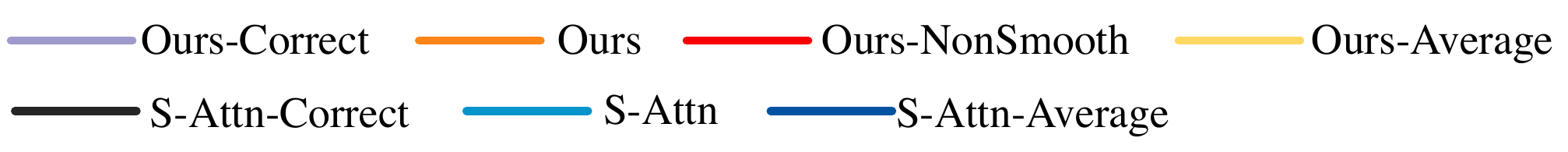}
    \vspace{-25pt}
    \caption{The left figures illustrate sample trajectories from the ground-truth policies for the major and minor cases of the Double Merge scenario. In the minor case, the left green vehicle is ahead of the right brown vehicle at initialization, and vice versa in the major case. The right figures are the Average Displacement Error (ADE) and the Final Displacement Error (FDE). We show the $p$-values that measure the statistical significance in the difference between the results of Ours and S-Attn.}
    \label{fig:interjunction_result}
    \vspace{-10pt}
\end{figure}

\smallskip
\noindent\textbf{Double Merge.} We define a common driving scenario illustrated in Fig.~\ref{fig:interjunction_result}~(left): two vehicles in two lanes (green and brown) want to change to each other's lane, which may cause a collision. We add $20$ other vehicles (white) that do not influence these two vehicles to make the scene more complex, requiring the two vehicles to attend each other instead of these other $20$ vehicles. We show sample trajectories, where the dot sizes represent progression in time.

We generate a trajectory dataset by using a hand-designed ground-truth policy for each vehicle. The main vehicle whose initial location is behind the other main vehicle (e.g. the brown car in Fig.~\ref{fig:interjunction_result} Minor) waits until the other (e.g., the green car) finishes its lane change, and then starts its own lane change. The other vehicles drive straight with a constant speed on the outer two lanes. Hence, the policy creates two symmetric situations where either of the two main vehicles is initialized in the front (Minor and Major cases in Fig.~\ref{fig:interjunction_result}).

We aim to show the regularization reduces the sample complexity, and improves performance especially for the rare events where data is limited. For this, we limit the data for one of these two driving scenarios.
Specifically, we assume the case the green vehicle starts behind the brown vehicle is a \emph{major} case with significant amount of data, while the other scenario as the \emph{minor} case with only limited data.
We collect $50$ trajectories from the major case for training. For the minor case, we vary the size of the dataset by collecting $30\%$ and $80\%$ of the number of major case trajectories. We further split $20\%$ of the entire training set for validation. For testing, we independently sample $50$ trajectories for each case and separately compute the prediction error on the test set for each situation. 

\begin{figure}
    \centering
    \includegraphics[width=.235\textwidth]{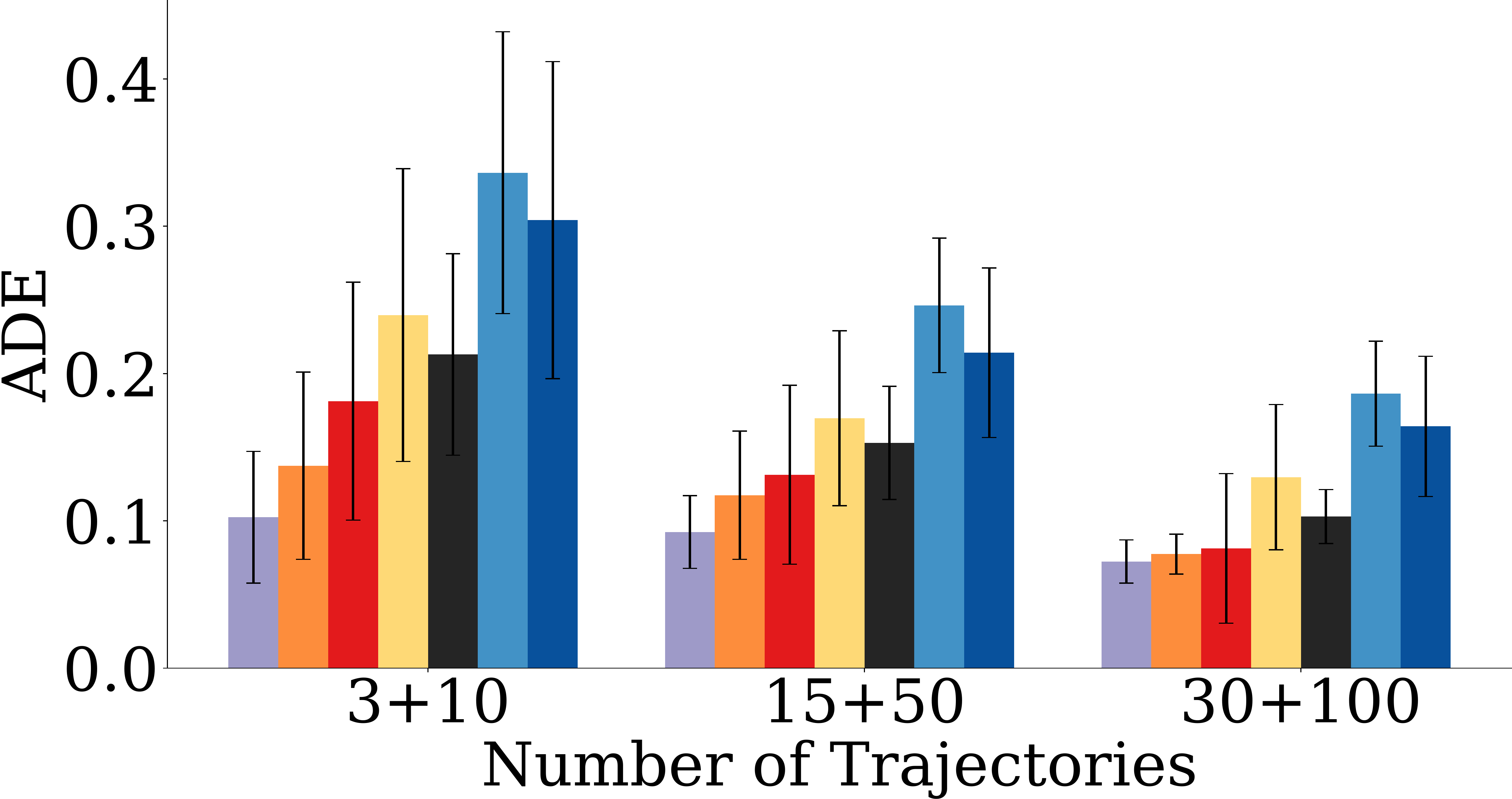}
    \includegraphics[width=.235\textwidth]{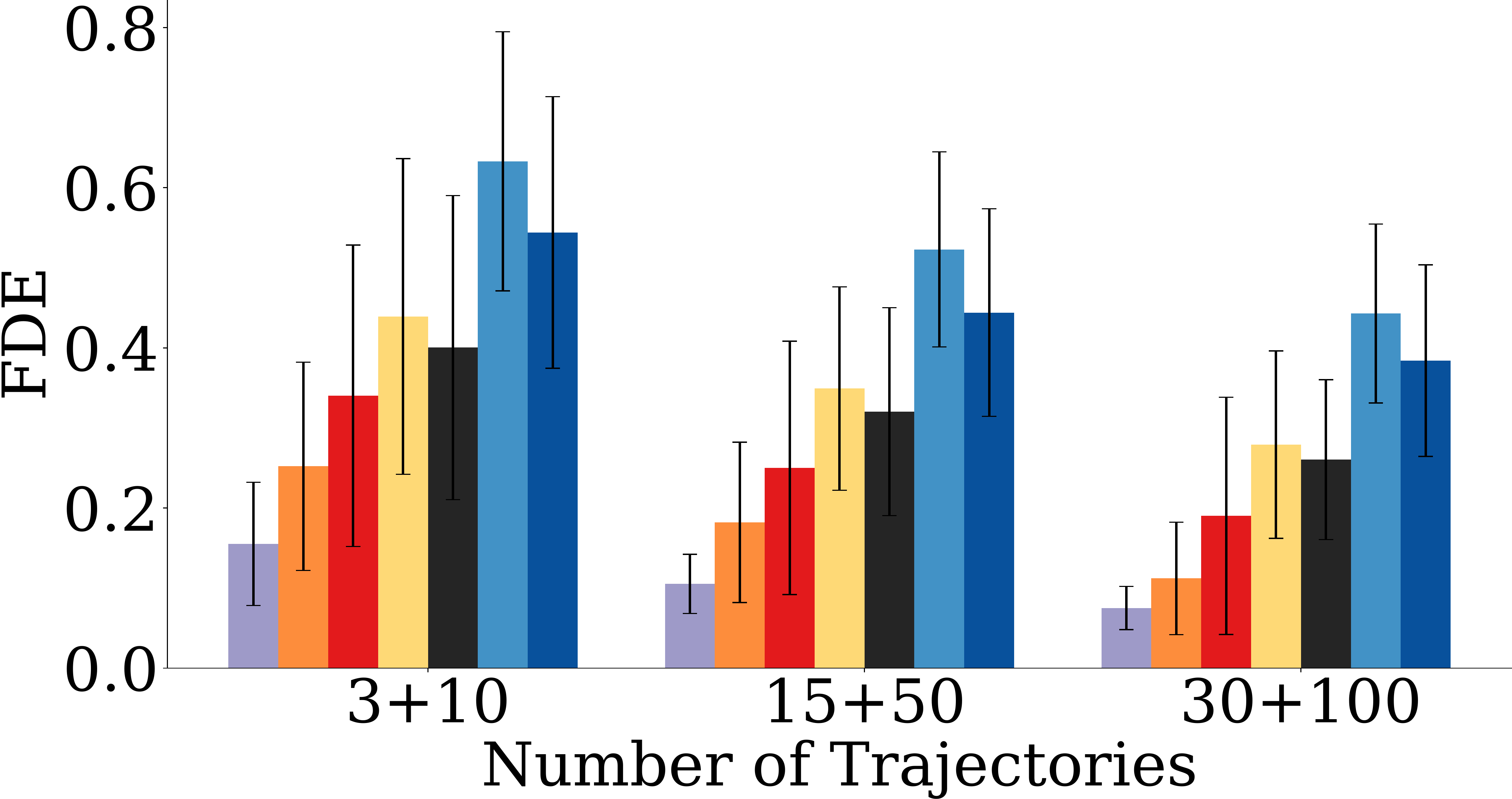}
    \vspace{-10pt}
    \caption{The test error under varying number of training trajectories. $x$-axes show the number of trajectories for minor$+$major cases. Our method is consistently better with a larger margin under smaller datasets.}
    \label{fig:valid_error}
    \vspace{-20pt}
\end{figure}

\smallskip
\noindent\textbf{Halting Car.} In this scenario, illustrated in Fig.~\ref{fig:haltingcar_result}, two main vehicles are initialized in the same lane while one vehicle (the leader, brown) is in front of the other (the follower, green). We create two scenarios: `Go', where the leader vehicle drives with a constant speed, and `Stop', where the leader vehicle suddenly stops and then accelerates back. The follower vehicle aims to follow the leader vehicle safely, and thus needs to react to the potential slowing down behavior of the leader. We also add $20$ other cars to increase complexity.

For the ground-truth policies, we let the leader vehicle slow down to stopping and then accelerate in the `Stop' case. The follower vehicle slows down or accelerates depending on its distance to the leader vehicle. In the `Go' case, both the leader and the follower vehicles drive with a constant speed. In both cases, the other vehicles drive straight with a constant speed on the outer lanes. The scenario aims to simulate a near-accident driving situation.

We simulate the rare events by using either setting (`Stop'/`Go') as the major case and the other as the minor case. We collect $50$ trajectories for the major case for training, and collect $30\%$ of the major case trajectories for the minor case. The training/validation split and the number of testing trajectories are the same as the double merge scenario.

\smallskip
\noindent\textbf{INTERACTION.} The INTERACTION dataset~\cite{interactiondataset} contains naturalistic motions of various traffic participants in 3 categories of highly interactive driving scenarios including Merging, Intersection, and Roundabout from 12 locations. The trajectories of all the agents in a location are recorded for prediction during a certain time horizon. For every $100$ steps, we use the first $40$ steps as observed steps and predict the next $60$ steps. For each dataset, we split all sequences into $80\%$ for training, $10\%$ for validation and $10\%$ for testing. The total time of video recorded is about 1000 minutes, which shows the performance of our method on a large-scale dataset.

\subsection{Implementation Details}
We use a one-layer LSTM~\cite{hochreiter1997long} for all RNN networks and use a one-layer fully-connected network for all other networks in our architecture. We use Adam optimizer~\cite{kingma2014adam} and tune the hyper-parameters; the learning rate, the threshold $\tau$, and the trade-off parameters $\beta_1$ and $\beta_2$, by cross-validation on the validation set. We set the learning rate to be $0.001$, $\tau$ as $0.001$ and both $\beta_1$ and $\beta_2$ to be $0.01$, which work well for all of our datasets. We use a batch size of $8$ and train the model on synthetic datasets with 200 epochs (only about 10 iterations per epoch) and on the Interaction dataset with 10 epochs. We run the experiments for 10 times, and report the mean and standard deviation.

\begin{figure}[t]
    \centering
     \includegraphics[width=\columnwidth]{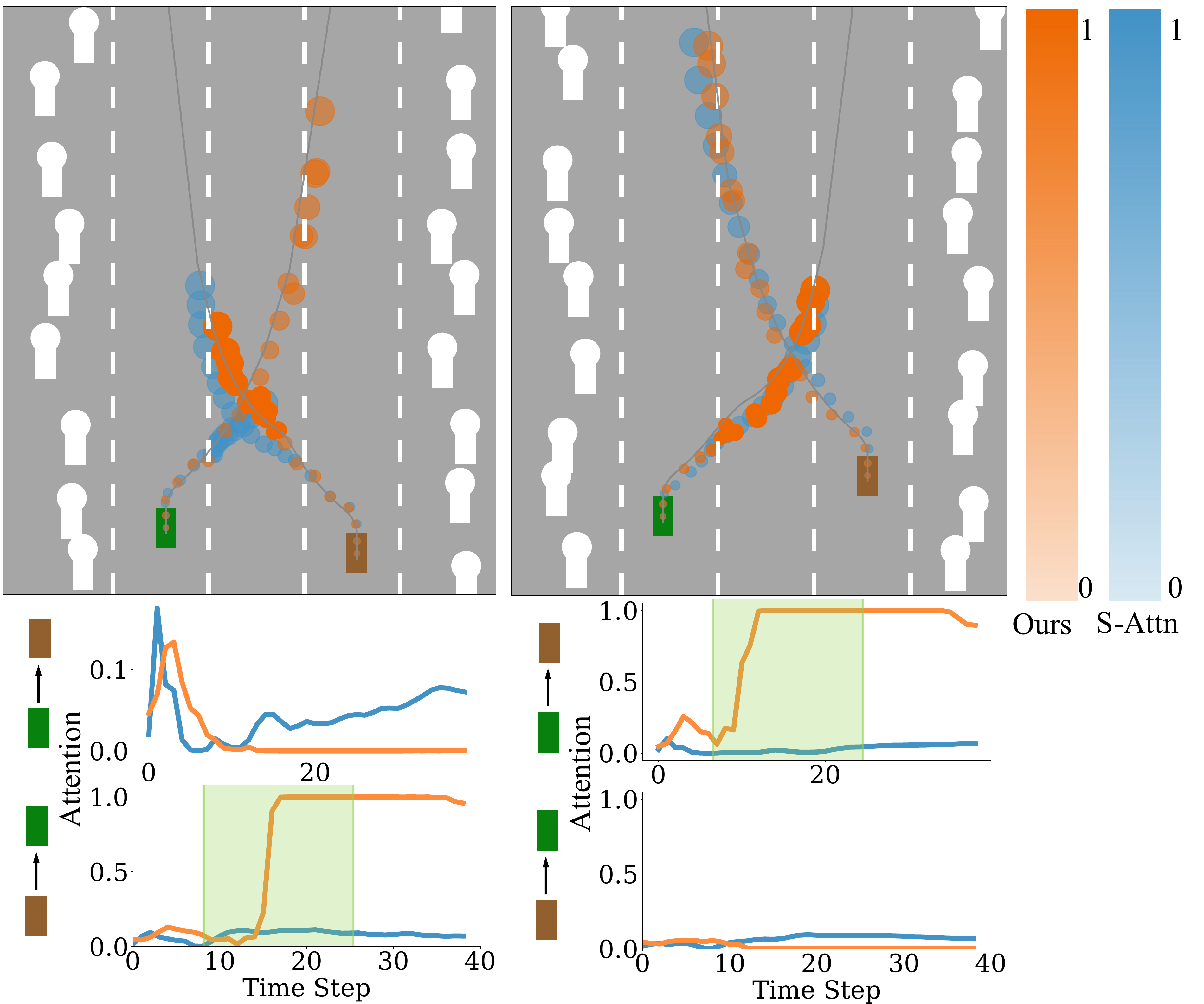}
     \vspace{-25pt}
    \caption{The top figures show the predicted trajectory. We show the trajectory of S-Attn with blue and our approach with orange. The darker the color at a step, the higher the attention paid on the correct vehicle (the green vehicle and the brown vehicle should pay attention to each other). The bottom plots show the attention at each step. The shaded area highlights the time period that requires high attention, meaning the vehicle behind is close to the center lane and needs to wait for the other vehicle ahead to pass first.}
    \label{fig:attention}
    \vspace{-15pt}
\end{figure}

\subsection{Results for Double Merge}
\smallskip
\noindent \textbf{Prediction Error.} We report the Average Displacement Error (ADE, mean error over the predicted steps) and the Final Displacement Error (FDE, the error in the final step). We compare our method with the (i) Social Attention (S-Attn)~\cite{vemula2018social}, which learns attention but does not have the sequence loss and the smoothness regularization, and (ii) the average attention (denoted by `Average'), which distributes the attention to all vehicles equally. We use the correct attention (denoted by `Correct') as the oracle: the two main vehicles fully attend to each other in the whole sequence and ignore other vehicles. As shown in Fig.~\ref{fig:interjunction_result}, in both the major and the minor cases, our approach consistently outperforms S-Attn with the same attention model, which demonstrates the importance of using the sequence loss. Ours outperforms Ours-NonSmooth and Ours-Average and performs comparably to Ours-Correct, which indicates that smooth attention regularization helps learn a more accurate attention. We report the $p$-values (two-sample $t$-test) in Fig.~\ref{fig:interjunction_result} for statistical significance of the difference between our method and S-Attn. We observe statistically significant performance gain ($p<0.01$) in both ADE and FDE for all the ratios in both major and minor cases.

We further investigate the test error with different number of trajectories, where we keep the ratio of the minor case to the major case as $30\%$ and change the total number of training trajectories. We test three settings (minor$ + $major): $3+10$, $15+30$ and $30+100$. As shown in Fig.~\ref{fig:valid_error}, Ours outperforms other methods with larger margin especially with less data, i.e., the margin between Ours and Ours-NonSmooth is larger for `$3+10$' than for `$30+100$'. This demonstrates the efficacy of smoothness regularizer on reducing sample complexity, which is crucial for driving in rare events, e.g., near-accident scenarios~\cite{cao2020reinforcement}.

\begin{figure}[t]
    \centering
    \includegraphics[width=\columnwidth]{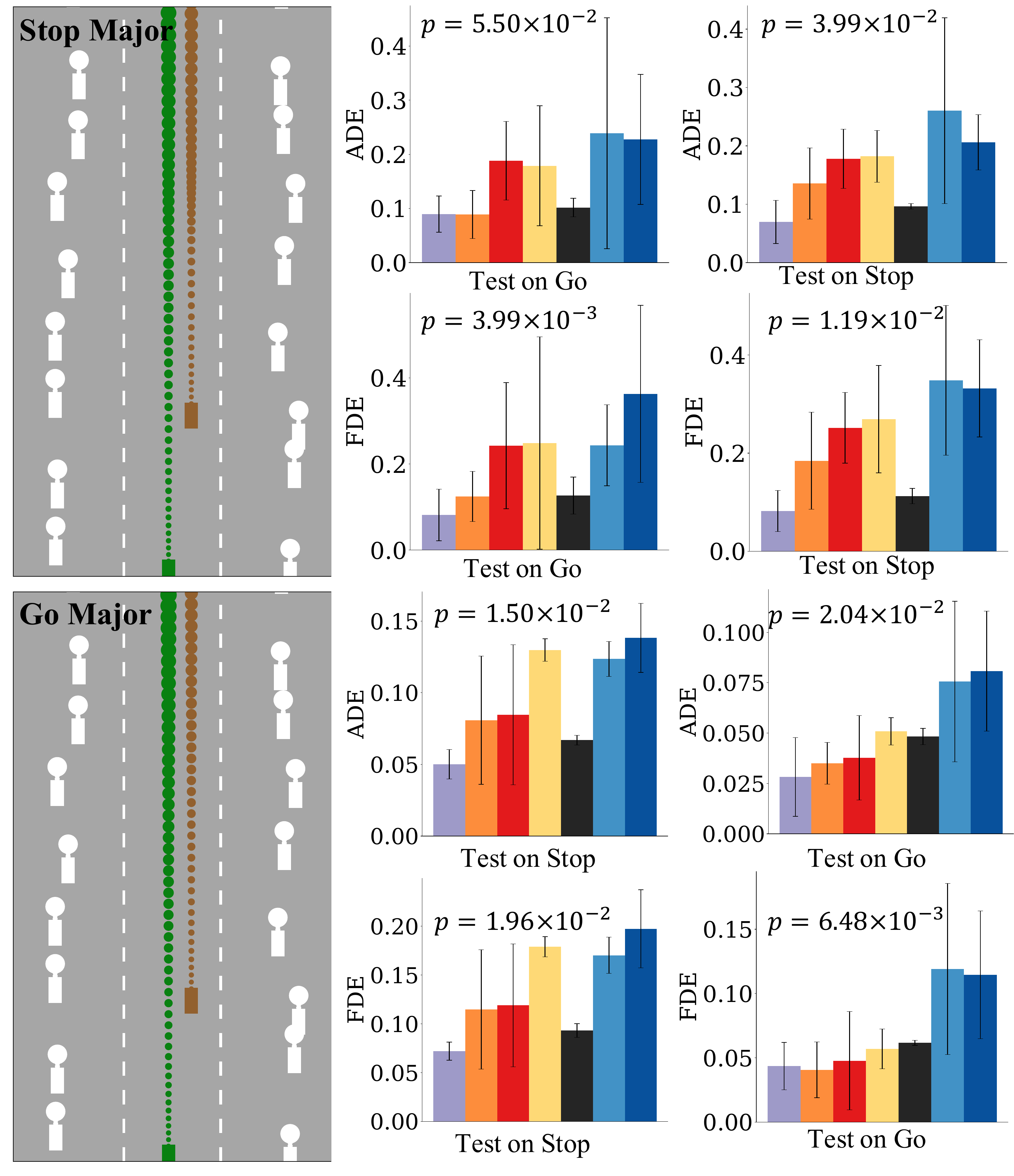}
    \includegraphics[width=\columnwidth]{figs/legend1.pdf}
    \vspace{-25pt}
    \caption{The left figures illustrate sample trajectories from the ground-truth policies for the major and minor cases of the Halting Car scenario. In the `Stop' case, the brown vehicle slows down to stopping, whereas it drives with a constant speed in the `Go' case. The right figures are the Average Displacement Error (ADE) and the Final Displacement Error (FDE) on the test set for these two cases. We show the $p$-values that measure the statistical significance in the difference between the results of Ours and S-Attn.}
    \label{fig:haltingcar_result}
    \vspace{-15pt}
\end{figure}

\smallskip
\noindent \textbf{Trajectory and Attention.} 
We show the trajectories and the attention for S-Attn and our approach in both minor and major cases in Fig.~\ref{fig:attention}. We show the true trajectories with the gray lines, and the predictions with dots. The sizes of the dots demonstrate progression in time, and the color intensities demonstrate attending to the correct vehicle. It can be seen that this intensity is larger when using our method (orange), compared to social attention (blue). In the bar plots in Fig.~\ref{fig:attention}, we demonstrate the predicted attention of the green vehicle to the brown vehicle (and vice versa) in both cases using our method (orange), and social attention (blue). The highlighted regions highlight the merging period. Our approach has a clear change in its value and correctly estimates the true attention in both major and minor cases compared to S-Attn.

\subsection{Results for Halting Car}
We use either `Stop' or `Go' as the major case and report ADE and FDE in both cases. As shown in Fig.~\ref{fig:haltingcar_result}, in both `Stop' major and the `Go' major cases, the results show a similar trend as the double merge scenario, which further demonstrates the efficacy of the smooth attention regularization and the sequence loss. The $p$-values in Fig.~\ref{fig:interjunction_result} demonstrate the statistically significant performance gain ($p<0.01$) in both ADE and FDE for all the ratios in both major and minor cases.

\begin{figure}
    \centering
    \subfigure{\includegraphics[width=.95\columnwidth]{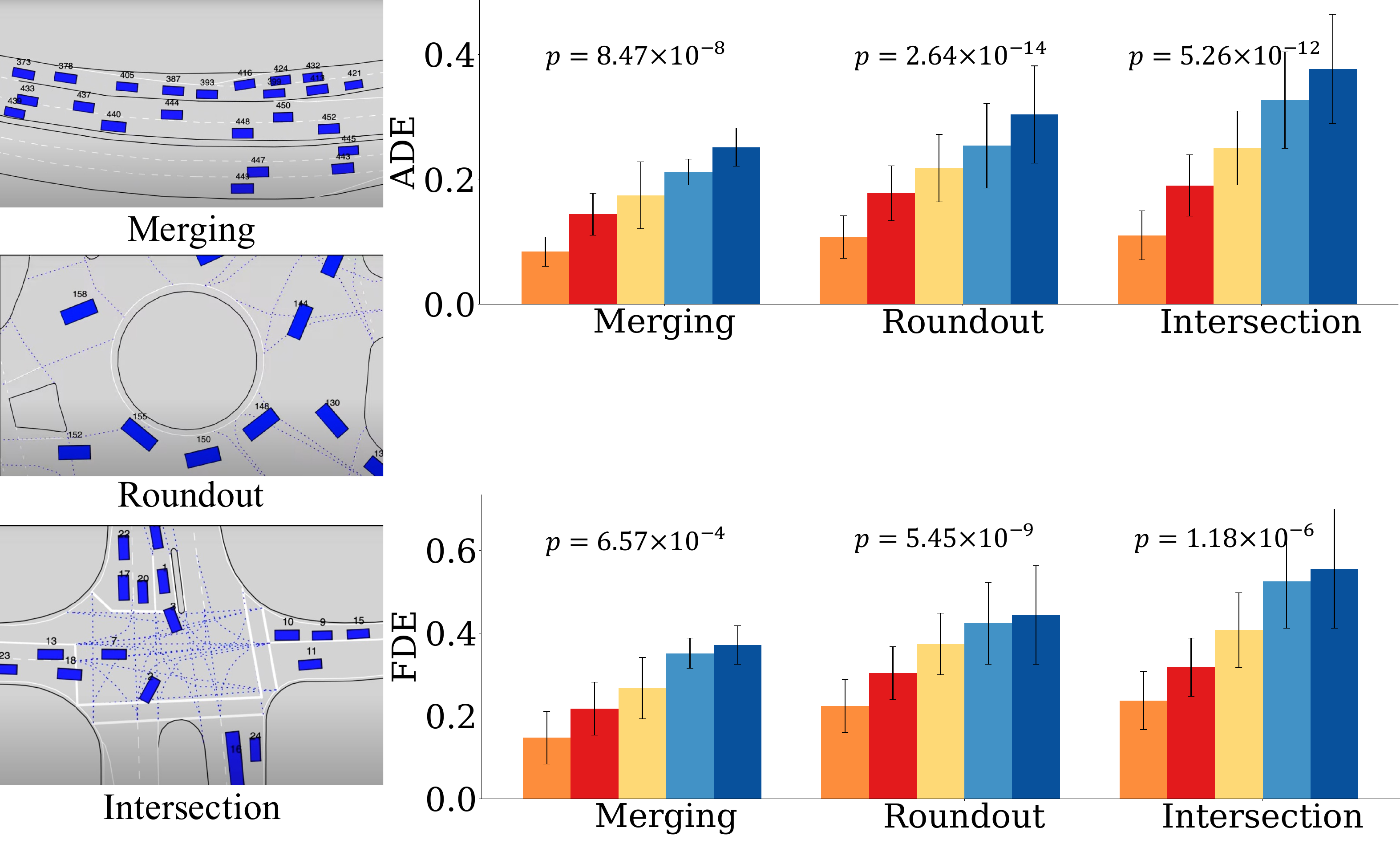}}\vskip -8px
    \subfigure{\includegraphics[width=.95\columnwidth]{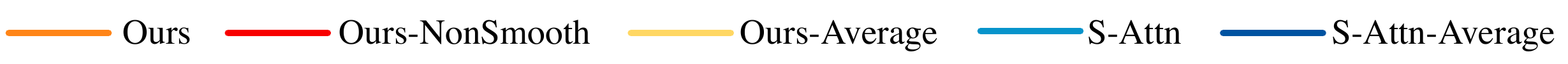}}
    \vspace{-10pt}
    \caption{The left figures illustrate different types of scenarios in the INTERACTION dataset. The right figures show Average Displacement Error (ADE) and Final Displacement Error (FDE) for different scenario types.}
    \label{fig:result_interaction}
    \vspace{-15pt}
\end{figure}
\subsection{Results for INTERACTION}
Fig.~\ref{fig:result_interaction} shows the results for the INTERACTION dataset. We compute the test error for each instance and report the average in each scenario type. We observe that our method outperforms all the other methods, which demonstrates the proposed smoothness regularization and the sequence loss improves the sequence prediction in realistic scenarios.

\section{Conclusion}
\noindent \textbf{Summary.} We propose a new approach for multi-agent trajectory prediction in complex driving scenarios. We propose smoothness attention prior motivated by cognitive evidence and realistic driving behavior. We design an RNN mixture architecture to model the interactions between agents and learn an attention module with a loss that consists of both one-step and sequence losses to penalize both short-term and long-term prediction errors, and a total variation loss on the attention model to impose the smoothness prior. Experimental results show the proposed approach outperforms the benchmarks on both synthetic and real driving scenarios. 

\noindent \textbf{Limitations and Future Work.} 
Our main focus is on how the attention prior affects the attention model but not the architecture to learn interactions. We use a complete directed graph to model the interactions between all agents, which may not be scalable to large number of agents. Also, we only consider learning a training model for one scenario but not for multiple scenarios. In the future, we plan to model interactions by sparse and/or dynamic graphs, and also use the scene maps as an input to handle multiple scenarios. The more scalable architectures may also enable conducting experiments on larger-scale real driving datasets where rare event naturally occur, e.g., Argoverse~\cite{chang2019argoverse}.

\section*{Acknowledgments}
We would like to thank NSF Awards 1953032 and 2125511, as well as Toyota Research Institute (TRI).

{\small
\bibliography{main.bib}

\begin{thebibliography}{10}
\providecommand{\url}[1]{#1}
\csname url@rmstyle\endcsname
\providecommand{\newblock}{\relax}
\providecommand{\bibinfo}[2]{#2}
\providecommand\BIBentrySTDinterwordspacing{\spaceskip=0pt\relax}
\providecommand\BIBentryALTinterwordstretchfactor{4}
\providecommand\BIBentryALTinterwordspacing{\spaceskip=\fontdimen2\font plus
\BIBentryALTinterwordstretchfactor\fontdimen3\font minus
  \fontdimen4\font\relax}
\providecommand\BIBforeignlanguage[2]{{%
\expandafter\ifx\csname l@#1\endcsname\relax
\typeout{** WARNING: IEEEtran.bst: No hyphenation pattern has been}%
\typeout{** loaded for the language `#1'. Using the pattern for}%
\typeout{** the default language instead.}%
\else
\language=\csname l@#1\endcsname
\fi
#2}}

\bibitem{lefevre2014survey}
S.~Lef{\`e}vre, D.~Vasquez, and C.~Laugier, ``A survey on motion prediction and
  risk assessment for intelligent vehicles,'' \emph{ROBOMECH journal}, vol.~1,
  no.~1, pp. 1--14, 2014.

\bibitem{schwarting2019social}
W.~Schwarting, A.~Pierson, J.~Alonso-Mora, S.~Karaman, and D.~Rus, ``Social
  behavior for autonomous vehicles,'' \emph{Proceedings of the National Academy
  of Sciences}, vol. 116, no.~50, pp. 24\,972--24\,978, 2019.

\bibitem{chang2019argoverse}
M.-F. Chang, J.~Lambert, P.~Sangkloy, J.~Singh, S.~Bak, A.~Hartnett, D.~Wang,
  P.~Carr, S.~Lucey, D.~Ramanan, \emph{et~al.}, ``Argoverse: {3D} tracking and
  forecasting with rich maps,'' in \emph{CVPR}, 2019, pp. 8748--8757.

\bibitem{houston2020one}
J.~Houston, G.~Zuidhof, L.~Bergamini, Y.~Ye, A.~Jain, S.~Omari, V.~Iglovikov,
  and P.~Ondruska, ``One thousand and one hours: Self-driving motion prediction
  dataset,'' in \emph{CoRL}, 2020.

\bibitem{bhattacharyya2018multi}
R.~P. Bhattacharyya, D.~J. Phillips, B.~Wulfe, J.~Morton, A.~Kuefler, and M.~J.
  Kochenderfer, ``Multi-agent imitation learning for driving simulation,'' in
  \emph{IROS}, 2018, pp. 1534--1539.

\bibitem{scalf2013competition}
P.~E. Scalf, A.~Torralbo, E.~Tapia, and D.~M. Beck, ``Competition explains
  limited attention and perceptual resources: implications for perceptual load
  and dilution theories,'' \emph{Frontiers in psychology}, vol.~4, p. 243,
  2013.

\bibitem{treisman1969modelsofattention}
A.~Treisman, ``Strategies and models of selective attention,''
  \emph{Psychological review}, vol.~76, no.~3, p. 282—299, 1969.

\bibitem{vemula2018social}
A.~Vemula, K.~Muelling, and J.~Oh, ``Social attention: Modeling attention in
  human crowds,'' in \emph{ICRA}, 2018, pp. 1--7.

\bibitem{gupta2018social}
A.~Gupta, J.~Johnson, L.~Fei-Fei, S.~Savarese, and A.~Alahi, ``Social {GAN}:
  Socially acceptable trajectories with generative adversarial networks,'' in
  \emph{CVPR}, 2018, pp. 2255--2264.

\bibitem{casas2020importance}
S.~Casas, C.~Gulino, S.~Suo, and R.~Urtasun, ``The importance of prior
  knowledge in precise multimodal prediction,'' in \emph{2020 IEEE/RSJ
  International Conference on Intelligent Robots and Systems (IROS)}.\hskip 1em
  plus 0.5em minus 0.4em\relax IEEE, 2020, pp. 2295--2302.

\bibitem{alahi2016social}
A.~Alahi, K.~Goel, V.~Ramanathan, A.~Robicquet, L.~Fei-Fei, and S.~Savarese,
  ``Social lstm: Human trajectory prediction in crowded spaces,'' in
  \emph{CVPR}, 2016, pp. 961--971.

\bibitem{zhang2018attention}
J.~Zhang, Y.~Zhao, H.~Li, and C.~Zong, ``Attention with sparsity regularization
  for neural machine translation and summarization,'' \emph{IEEE/ACM
  Transactions on Audio, Speech, and Language Processing}, vol.~27, no.~3, pp.
  507--518, 2018.

\bibitem{swettenham1998frequency}
J.~Swettenham, S.~Baron-Cohen, T.~Charman, A.~Cox, G.~Baird, A.~Drew, L.~Rees,
  and S.~Wheelwright, ``The frequency and distribution of spontaneous attention
  shifts between social and nonsocial stimuli in autistic, typically
  developing, and nonautistic developmentally delayed infants,'' \emph{Journal
  of child Psychology and Psychiatry}, vol.~39, no.~5, pp. 747--753, 1998.

\bibitem{helbing1995social}
D.~Helbing and P.~Molnar, ``Social force model for pedestrian dynamics,''
  \emph{Physical review E}, vol.~51, no.~5, p. 4282, 1995.

\bibitem{kim2011gaussian}
K.~Kim, D.~Lee, and I.~Essa, ``Gaussian process regression flow for analysis of
  motion trajectories,'' in \emph{ICCV}, 2011, pp. 1164--1171.

\bibitem{li2019generic}
J.~Li, W.~Zhan, Y.~Hu, and M.~Tomizuka, ``Generic tracking and probabilistic
  prediction framework and its application in autonomous driving,''
  \emph{T-ITS}, vol.~21, no.~9, pp. 3634--3649, 2019.

\bibitem{kasper2012object}
D.~Kasper, G.~Weidl, T.~Dang, G.~Breuel, A.~Tamke, A.~Wedel, and W.~Rosenstiel,
  ``Object-oriented bayesian networks for detection of lane change maneuvers,''
  \emph{ITS Magazine}, vol.~4, no.~3, pp. 19--31, 2012.

\bibitem{sun2018probabilistic}
L.~Sun, W.~Zhan, and M.~Tomizuka, ``Probabilistic prediction of interactive
  driving behavior via hierarchical inverse reinforcement learning,'' in
  \emph{ITSC}.\hskip 1em plus 0.5em minus 0.4em\relax IEEE, 2018, pp.
  2111--2117.

\bibitem{chandra2020forecasting}
R.~Chandra, T.~Guan, S.~Panuganti, T.~Mittal, U.~Bhattacharya, A.~Bera, and
  D.~Manocha, ``Forecasting trajectory and behavior of road-agents using
  spectral clustering in graph-lstms,'' \emph{RA-L}, vol.~5, no.~3, pp.
  4882--4890, 2020.

\bibitem{huang2020diversitygan}
X.~Huang, S.~G. McGill, J.~A. DeCastro, L.~Fletcher, J.~J. Leonard, B.~C.
  Williams, and G.~Rosman, ``Diversitygan: Diversity-aware vehicle motion
  prediction via latent semantic sampling,'' \emph{RA-L}, vol.~5, no.~4, pp.
  5089--5096, 2020.

\bibitem{pokle2019deep}
A.~Pokle, R.~Martin-Martin, P.~Goebel, V.~Chow, H.~M. Ewald, J.~Yang,
  W.~Zhenkai, A.~Sadeghian, D.~Sadigh, S.~Savarese, and M.~Vazquez, ``Deep
  local trajectory replanning and control for robot navigation,'' in
  \emph{International Conference on Robotics and Automation (ICRA)}, May 2019.

\bibitem{basu2019active}
C.~Basu, E.~Biyik, Z.~He, M.~Singhal, and D.~Sadigh, ``Active learning of
  reward dynamics from hierarchical queries,'' in \emph{IROS}, 2019.

\bibitem{kwon2020when}
M.~Kwon, E.~Biyik, A.~Talati, K.~Bhasin, D.~P. Losey, and D.~Sadigh, ``When
  humans aren't optimal: Robots that collaborate with risk-aware humans,'' in
  \emph{HRI}, 2020.

\bibitem{xie2020learning}
A.~Xie, D.~Losey, R.~Tolsma, C.~Finn, and D.~Sadigh, ``Learning latent
  representations to influence multi-agent interaction,'' in \emph{CoRL}, 2020.

\bibitem{zhu2020multi}
Z.~Zhu, E.~Biyik, and D.~Sadigh, ``Multi-agent safe planning with gaussian
  processes,'' in \emph{Proceedings of the IEEE/RSJ International Conference on
  Intelligent Robots and Systems (IROS)}, oct 2020.

\bibitem{sadigh2016information}
D.~Sadigh, S.~S. Sastry, S.~A. Seshia, and A.~Dragan, ``Information gathering
  actions over human internal state,'' in \emph{Proceedings of the {IEEE},
  /{RSJ}, International Conference on Intelligent Robots and Systems
  (IROS)}.\hskip 1em plus 0.5em minus 0.4em\relax IEEE, Oct. 2016, pp. 66--73.

\bibitem{sadigh2016planning}
D.~Sadigh, S.~S. Sastry, S.~A. Seshia, and A.~D. Dragan, ``Planning for
  autonomous cars that leverage effects on human actions,'' in
  \emph{Proceedings of Robotics: Science and Systems (RSS)}, June 2016.

\bibitem{sadigh2018planning}
D.~Sadigh, N.~Landolfi, S.~S. Sastry, S.~A. Seshia, and A.~D. Dragan,
  ``Planning for cars that coordinate with people: Leveraging effects on human
  actions for planning and active information gathering over human internal
  state,'' \emph{Autonomous Robots (AURO)}, vol.~42, no.~7, pp. 1405--1426,
  Oct. 2018.

\bibitem{sukhbaatar2016learning}
S.~Sukhbaatar, A.~Szlam, and R.~Fergus, ``Learning multiagent communication
  with backpropagation,'' in \emph{NeurIPS}, vol.~29, 2016.

\bibitem{shih2021critical}
A.~Shih, A.~Sawhney, J.~Kondic, S.~Ermon, and D.~Sadigh, ``On the critical role
  of conventions in adaptive human-ai collaboration,'' in \emph{ICLR}, 2021.

\bibitem{kipf2018neural}
T.~Kipf, E.~Fetaya, K.-C. Wang, M.~Welling, and R.~Zemel, ``Neural relational
  inference for interacting systems,'' in \emph{International Conference on
  Machine Learning}.\hskip 1em plus 0.5em minus 0.4em\relax PMLR, 2018, pp.
  2688--2697.

\bibitem{bohmer2020deep}
W.~B{\"o}hmer, V.~Kurin, and S.~Whiteson, ``Deep coordination graphs,'' in
  \emph{ICML}, 2020, pp. 980--991.

\bibitem{casas2020implicit}
S.~Casas, C.~Gulino, S.~Suo, K.~Luo, R.~Liao, and R.~Urtasun, ``Implicit latent
  variable model for scene-consistent motion forecasting,'' in \emph{Computer
  Vision--ECCV 2020: 16th European Conference, Glasgow, UK, August 23--28,
  2020, Proceedings, Part XXIII 16}.\hskip 1em plus 0.5em minus 0.4em\relax
  Springer, 2020, pp. 624--641.

\bibitem{salzmann2020trajectron++}
T.~Salzmann, B.~Ivanovic, P.~Chakravarty, and M.~Pavone, ``Trajectron++:
  Dynamically-feasible trajectory forecasting with heterogeneous data,'' in
  \emph{Computer Vision--ECCV 2020: 16th European Conference, Glasgow, UK,
  August 23--28, 2020, Proceedings, Part XVIII 16}.\hskip 1em plus 0.5em minus
  0.4em\relax Springer, 2020, pp. 683--700.

\bibitem{li2020evolvegraph}
J.~Li, F.~Yang, M.~Tomizuka, and C.~Choi, ``Evolvegraph: Multi-agent trajectory
  prediction with dynamic relational reasoning,'' in \emph{NeurIPS}, 2020.

\bibitem{messaoud2020attention}
K.~Messaoud, I.~Yahiaoui, A.~Verroust, and F.~Nashashibi, ``Attention based
  vehicle trajectory prediction,'' \emph{IV}, 2020.

\bibitem{bahdanau2015neural}
D.~Bahdanau, K.~Cho, and Y.~Bengio, ``Neural machine translation by jointly
  learning to align and translate,'' in \emph{ICLR}, 2015.

\bibitem{luong2015effective}
M.-T. Luong, H.~Pham, and C.~D. Manning, ``Effective approaches to
  attention-based neural machine translation,'' in \emph{EMNLP}, 2015, pp.
  1412--1421.

\bibitem{martins2020sparse}
A.~Martins, A.~Farinhas, M.~Treviso, V.~Niculae, P.~Aguiar, and M.~Figueiredo,
  ``Sparse and continuous attention mechanisms,'' in \emph{NeurIPS}, vol.~33,
  2020, pp. 20\,989--21\,001.

\bibitem{zhou2019discriminative}
S.~Zhou, F.~Wang, Z.~Huang, and J.~Wang, ``Discriminative feature learning with
  consistent attention regularization for person re-identification,'' in
  \emph{ICCV}, 2019, pp. 8040--8049.

\bibitem{niculae2017regularized}
V.~Niculae and M.~Blondel, ``A regularized framework for sparse and structured
  neural attention,'' in \emph{NeurIPS}, vol.~30, 2017.

\bibitem{li-etal-2018-multi-head}
J.~Li, Z.~Tu, B.~Yang, M.~R. Lyu, and T.~Zhang, ``Multi-head attention with
  disagreement regularization,'' in \emph{EMNLP}, 2018, pp. 2897--2903.

\bibitem{wolfe2000attention}
J.~M. Wolfe, G.~A. Alvarez, and T.~S. Horowitz, ``Attention is fast but
  volition is slow,'' \emph{Nature}, vol. 406, no. 6797, pp. 691--691, 2000.

\bibitem{rudin1992nonlinear}
L.~I. Rudin, S.~Osher, and E.~Fatemi, ``Nonlinear total variation based noise
  removal algorithms,'' \emph{Physica D: nonlinear phenomena}, vol.~60, no.
  1-4, pp. 259--268, 1992.

\bibitem{bresson2008fast}
X.~Bresson and T.~F. Chan, ``Fast dual minimization of the vectorial total
  variation norm and applications to color image processing,'' \emph{Inverse
  Problems \& Imaging}, vol.~2, no.~4, p. 455, 2008.

\bibitem{cao2020reinforcement}
Z.~Cao, E.~Biyik, W.~Z. Wang, A.~Raventos, A.~Gaidon, G.~Rosman, and D.~Sadigh,
  ``Reinforcement learning based control of imitative policies for
  near-accident driving,'' in \emph{Proceedings of Robotics: Science and
  Systems (RSS)}, 7 2020.

\bibitem{interactiondataset}
W.~Zhan, L.~Sun, D.~Wang, H.~Shi, A.~Clausse, M.~Naumann, J.~K\"ummerle,
  H.~K\"onigshof, C.~Stiller, A.~de~La~Fortelle, and M.~Tomizuka,
  ``{INTERACTION} {Dataset}: {An} {INTERnational}, {Adversarial} and
  {Cooperative} {moTION} {Dataset} in {Interactive} {Driving} {Scenarios} with
  {Semantic} {Maps},'' \emph{arXiv:1910.03088 [cs, eess]}, 2019.

\bibitem{hochreiter1997long}
S.~Hochreiter and J.~Schmidhuber, ``Long short-term memory,'' \emph{Neural
  computation}, vol.~9, no.~8, pp. 1735--1780, 1997.

\bibitem{kingma2014adam}
D.~P. Kingma and J.~Ba, ``Adam: A method for stochastic optimization,''
  \emph{arXiv preprint arXiv:1412.6980}, 2014.

\end{thebibliography}
\bibliographystyle{IEEEtran}
}

\end{document}